\documentclass[conf]{IEEEtran}
\usepackage{algorithm,algpseudocode}
\usepackage{amsmath}
\usepackage{amssymb}
\usepackage{mathtools}

\usepackage{cite}
\def\BibTeX{{\rm B\kern-.05em{\sc i\kern-.025em b}\kern-.08em
    T\kern-.1667em\lower.7ex\hbox{E}\kern-.125emX}}
\usepackage[bookmarksnumbered=true]{hyperref}

\hypersetup{
     colorlinks = true,
     linkcolor = blue,
     anchorcolor = blue,
     citecolor = blue,
     filecolor = blue,
     urlcolor = blue
     }

\usepackage{svg}

\usepackage{wrapfig}
\usepackage{caption}
\usepackage{subcaption}
\usepackage{booktabs}

\usepackage{xcolor}
\begin{document}

\title{Examining and Mitigating the Impact of Crossbar Non-idealities for Accurate Implementation of Sparse Deep Neural Networks}

\author{
\IEEEauthorblockN{Abhiroop Bhattacharjee$^{1}$, Lakshya Bhatnagar$^{2}$, and Priyadarshini Panda$^{1}$} \\
\IEEEauthorblockA{
$^{1}$Department of Electrical Engineering, Yale University, USA\\ 
$^{2}$Indian Institute of Technology, Delhi, India\\
\{abhiroop.bhattacharjee, priya.panda\}@yale.edu 
}
}


\IEEEaftertitletext{\vspace{-2.5\baselineskip}}
\maketitle


\begin{abstract}

Recently several structured pruning techniques have been introduced for energy-efficient implementation of Deep Neural Networks (DNNs) with lesser number of crossbars. Although, these techniques have claimed to preserve the accuracy of the sparse DNNs on crossbars, none have studied the impact of the inexorable crossbar non-idealities on the actual performance of the pruned networks. To this end, we perform a comprehensive study to show how highly sparse DNNs, that result in significant crossbar-compression-rate, can lead to severe accuracy losses compared to unpruned DNNs mapped onto non-ideal crossbars. We perform experiments with multiple structured-pruning approaches (such as, C/F pruning,  XCS and XRS) on VGG11 and VGG16 DNNs with benchmark datasets (CIFAR10 and CIFAR100). We propose two mitigation approaches - Crossbar-column rearrangement and Weight-Constrained-Training (WCT) - that can be integrated with the crossbar-mapping of the sparse DNNs to minimize accuracy losses incurred by the pruned models. These help in mitigating non-idealities by increasing the proportion of low conductance synapses on crossbars, thereby improving their computational accuracies.
\end{abstract}

\begin{IEEEkeywords}

Structured-pruning, crossbars, non-idealities, crossbar column-rearrangement, weight-constrained-training
\vspace{-5mm}
\end{IEEEkeywords}

\IEEEpeerreviewmaketitle

\section{Introduction}
\label{sec:intro}

The previous decade has seen the rise of Deep Neural Networks (DNNs) to solve various real world problems. To this end, memristive crossbar architectures have received significant attention to realize DNNs in an analog manner on hardware with high degree of parallelism, great compactness and energy-efficiency \cite{chi2016prime, jain2020rxnn}. Several Non-Volatile-Memory (NVM) devices such as, Resistive RAM (ReRAM), Phase Change Memory (PCM) and Spintronic devices have been explored as synapses to implement DNNs on crossbars \cite{chakraborty2020pathways}. 

In the recent years, several crossbar-aware pruning techniques have been devised that yield sparse DNN models \cite{wen2016learning, snrram, XCS, XRS}. Owing to their high sparsity, these models require significantly lower number of crossbars to be mapped, thereby introducing hardware resource-efficiency not only in terms of crossbars but also peripheral circuits interfacing the crossbars. Pruning algorithms such as, \cite{wen2016learning, snrram, XCS, XRS}, produce structured sparsity in DNNs that fit into crossbars as dense weight matrices \cite{pimprune}. These structured pruning algorithms claim to preserve the accuracy of the pruned DNNs, after implementation on crossbars, with minimal or no noticeable loss, while bringing in high energy- and area-efficiencies.  
However, analog crossbars possess several non-idealities such as, interconnect parasitics, non-linearities/variations in the synapses, etc. \cite{jain2020rxnn, bhattacharjee2021efficiency, bhattacharjee2021neat}. These non-idealities result in imprecise dot-product currents in the crossbars, leading to performance (accuracy) degradation on mapping DNNs. Many works have modelled these non-idealities and studied their repercussions on the performance and robustness of crossbar-mapped DNNs, and have suggested mitigation techniques \cite{chakraborty2020geniex, jain2020rxnn, bhattacharjee2021efficiency, bhattacharjee2021neat, liu2014reduction}. Though the above-mentioned crossbar-aware structured pruning algorithms have claimed to preserve the performance of the pruned DNNs, none of them have included the impact of the non-idealities during inference on crossbars. For a realistic hardware evaluation of the performance of increased structured sparsity in DNNs mapped on crossbars, the inclusion of hardware non-idealities is critical. Thus, this work is designed to draw the focus of the research community towards a non-ideality aware evaluation of the various existing structured pruning algorithms and showing how increased sparsity can degrade the performance of DNNs on non-ideal crossbars. In the end, we also introduce two hardware-centric non-ideality mitigation strategies, namely crossbar-column rearrangement and \textit{Weight-Constrained-Training} (WCT), to help improve the performance of the sparse DNNs on crossbars. 

The key contributions of this work are as follows:
\begin{itemize}
    \item We find that DNNs with high degree of structured sparsity suffer from greater performance degradation during inference on non-ideal crossbars with respect to unpruned DNN models.  
    
    \item There exists a \textit{trade-off}: as structured sparsity increases in DNNs, their mappings onto non-ideal crossbars become more resource-efficient in terms of area and energy, however with loss of performance (inference accuracy). 
    
    \item We propose a hardware-friendly non-ideality mitigation strategy called crossbar-column rearrangement, that increases the feasibility of low conductance synapses in crossbars, thereby reducing the impact of non-idealities.
    
    \item Finally, we integrate the crossbar-aware pruning techniques with \textit{Weight-Constrained-Training} (WCT), and show its effectiveness in non-ideality mitigation in sparse DNNs, especially on larger crossbars.
\end{itemize}

\vspace{-4mm}

\section{Background and Motivation}
\label{sec:background}

\subsection{Analog crossbars and their non-idealities}
\label{sec: xbar and ni}
\vspace{-1mm}

Memristive crossbars are used to realize Multiply-and-Accumulate (MAC) operations on hardware in an analog manner. Crossbars receive the input activations of a DNN as analog voltages and produce currents analogous to the outputs of MAC operations. Ideally, the MAC operations occur using Ohm's Law and Kirchoff's current law with the interaction of the input voltages and the memristive conductances for the synapses (programmed between $G_{MIN}$ and $G_{MAX}$). However, the analog nature of the computation leads to various non-idealities, such as, circuit-level interconnect parasitics and synaptic-level non-linearities or variations~\cite{bhattacharjee2021neat, bhattacharjee2021switchx, chakraborty2020geniex, jain2020rxnn}.


\begin{figure}[t]
    \centering
    \subfloat[]{
    \includegraphics[width=0.49\linewidth]{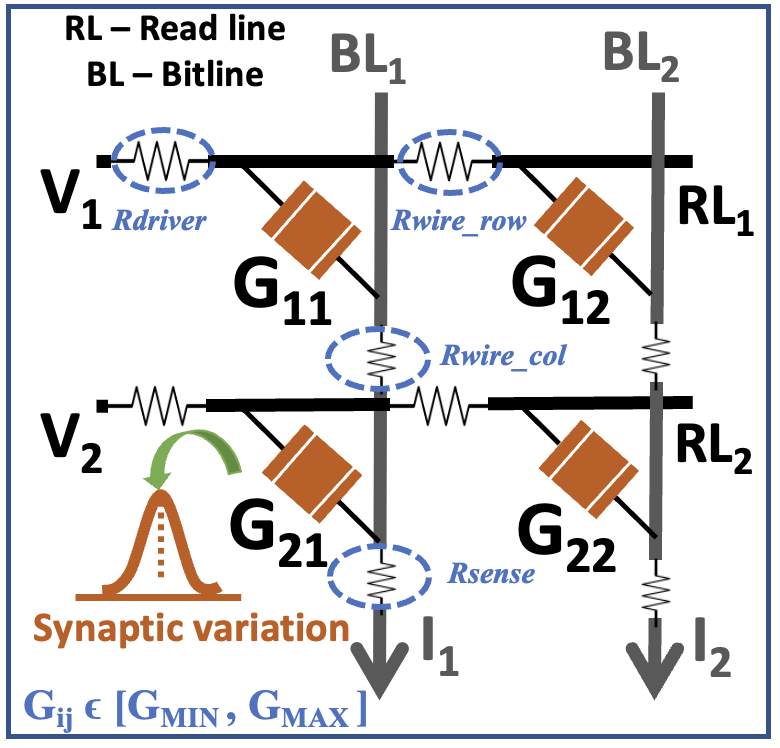}%
    }
    \subfloat[]{
    \includegraphics[width=0.43\linewidth]{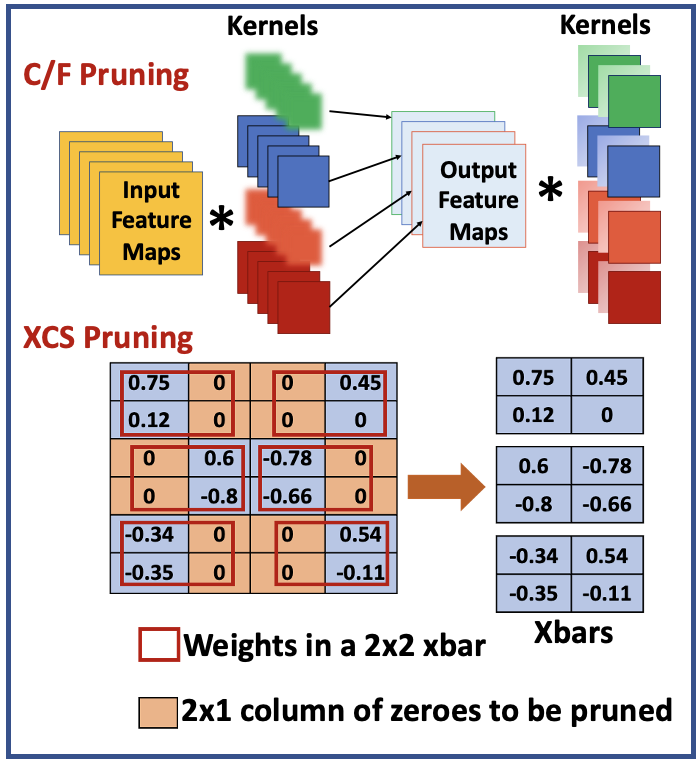}
    }
    \caption{(a) Non-ideal crossbar with input voltages $V_{i}$, synaptic conductances $G_{ij}$ and output currents $I_{j} = \sum_{i}^{}{G_{ij} * V_i}$. The interconnect and synaptic non-idealities, that lead to imprecise dot-product currents, are annotated; (b) \textbf{Top -} A representation of channel/filter pruning (C/F pruning). The blurred channels/filters correspond to DNN weights pruned in a structured manner. \textbf{Bottom -} A representation of XCS pruning}
    \label{xbar}
    \vspace{-7mm}
\end{figure}

\figurename{~\ref{xbar}}(a) describes the equivalent circuit for a memristive crossbar consisting of various circuit-level and device-level non-idealities, \textit{viz.} $Rdriver$, $Rwire\_row$, $Rwire\_col$ and $Rsense$ (interconnect parasitics), modelled as parasitic resistances and variations/non-linearities in the memristive synapses. The impact of these non-idealities can be incorporated by transforming the ideal memristive conductances $G_{ij(ideal)}$ to non-ideal conductances $G_{ij(non-ideal)}$. Consequently, the net output current sensed at the end of the crossbar-columns ($I_{non-ideal}$) deviates from its ideal value ($I_{ideal}$). This manifests as accuracy degradation for DNNs mapped onto crossbars. The relative deviation of $I_{non-ideal}$ from its ideal value is measured using \textit{non-ideality factor} (NF)~\cite{chakraborty2020geniex}. It is defined as $NF~=~(I_{ideal}-I_{non-ideal})/I_{ideal}$. NF is a direct measure of crossbar non-idealities, \textit{i.e.} increased non-idealities induce a greater value of NF, degrading the performance of the DNN mapped onto them.
\vspace{-6mm}

\subsection{Crossbar-aware structured pruning of DNNs}
\label{sec: pruning}
\vspace{-2mm}

In the recent years, there have been numerous works on structured pruning of DNNs, such as \textit{channel/filter pruning} or C/F pruning (see Fig. \ref{xbar}(b)-Top) wherein the unimportant filters and channels in a DNN (corresponding to rows and columns in the weight matrix of the DNN) are pruned to obtain a sparse 2D weight matrix \cite{wen2016learning, snrram}. These pruned models result in significant hardware savings in terms of reduced number of crossbars for mapping, thereby bringing in energy- and area-efficiency for DNN implementation. Likewise, other crossbar-aware pruning strategies include \textit{Crossbar Column Sparsity} (XCS) \cite{XCS} or \textit{Crossbar Row Sparsity} (XRS) \cite{XRS} (XCS shown in Fig. \ref{xbar}(b)-Bottom), that exploit fine-grained sparsity by respectively pruning columns or rows of weights within a crossbar \cite{pimprune}. Additionally, these works have claimed to  preserve  the inference accuracy  of  the  structure-pruned  networks  on  crossbars with minimal or no discernible performance loss  with respect to the unpruned ones. \textit{However, none of the previous works have accounted for the non-idealities inherent in crossbar arrays which raises concerns about the claimed performance of the highly pruned models in the real scenario.} 
\vspace{-8mm}
\section{Methodology and Evaluation Framework}
\label{sec:method}

    

  

In this work, we structure-prune DNNs at initialization with a given sparsity ratio ($s$) for each layer \cite{frankle2018lottery}, using the crossbar-aware techniques specified in Section \ref{sec: pruning}, and then train the respective pruned models on software. Traditionally, standard methods of obtaining sparse neural networks include training from scratch, followed by structured pruning and finally fine-tuning the pruned models \cite{pimprune}. However, structured pruning at initialization followed by training reduces training overheads since it requires only one round of training as opposed to the traditional approach requiring two rounds \cite{frankle2018lottery, malach2020proving}. 

\begin{figure}[htbp]
    \centering
    \includegraphics[width=0.9\linewidth]{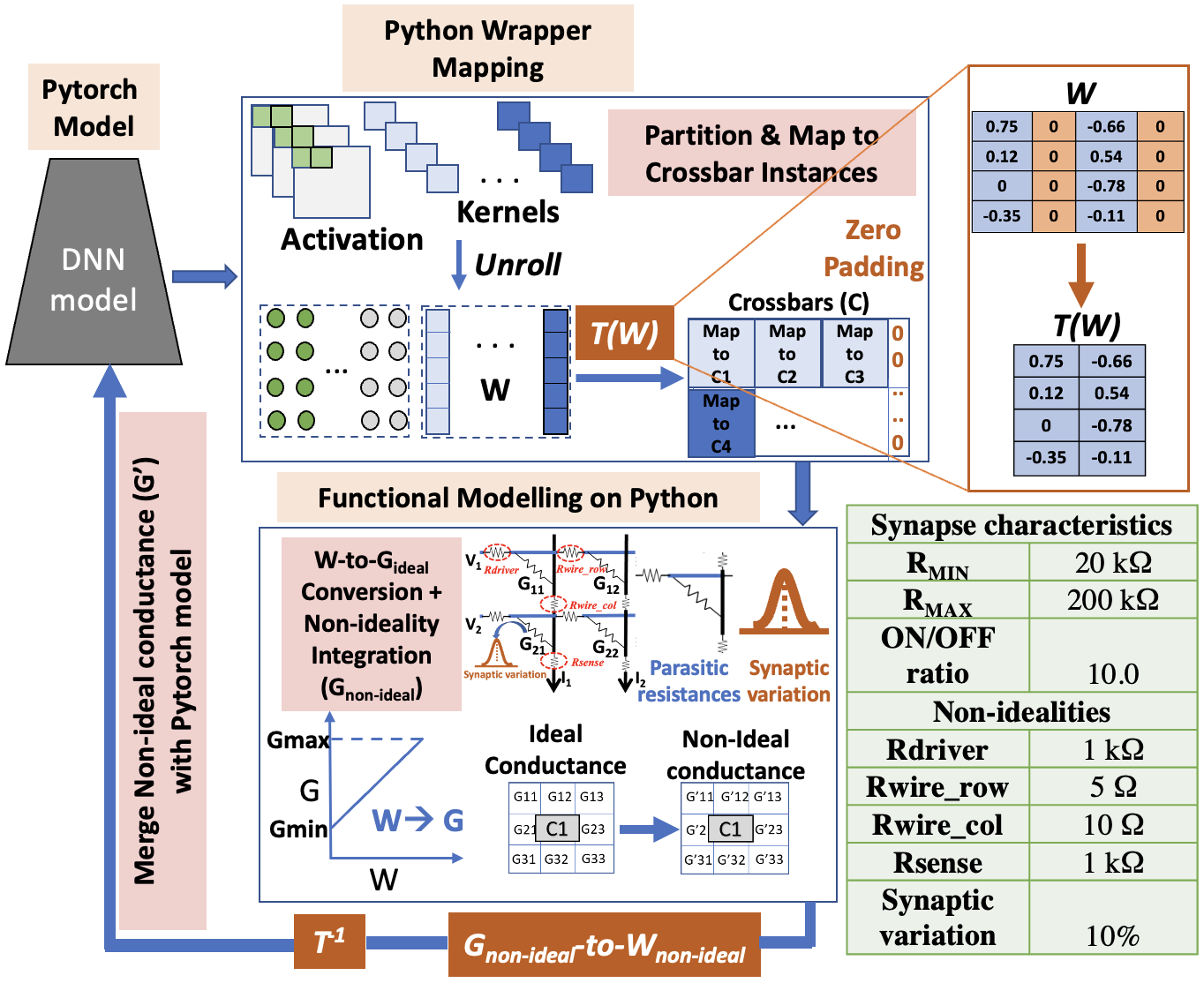}%
    \caption{Hardware evaluation framework in Python to map structure-pruned DNNs on non-ideal crossbars, followed by inference on the crossbar-mapped DNN models}
    \label{hardware_eval}
    \vspace{-4mm}
\end{figure}

Next, we  use  a  simulation framework  in Pytorch (see Fig. \ref{hardware_eval}) to map the trained DNNs onto non-ideal memristive crossbars  and  investigate  the  cumulative impact  of the  circuit  and  device-level  non-idealities  on  their performance during inference. In  the  platform,  a  Python  wrapper  is built  that  unrolls  each  and every  convolution  operation  in  the software  DNN  into  MAC  operations. This yields 2D weight matrices for each DNN layer which are to be partitioned into numerous crossbar instances of a given size (16$\times$16, 32$\times$32 or 64$\times$64). Before partitioning, based on the structured pruning approach we apply the following transformations $T$ on the sparse weight matrices $W$: 
\begin{enumerate}

\item \textbf{$T(W)$ for C/F pruning:} Here, for a given 2D weight matrix of a DNN layer, all the columns bearing zero values are eliminated. Further, we also eliminate rows of the weight matrix of the next DNN layer that interact with the output feature maps corresponding1 to the columns of zero values in the previous layer.

\item \textbf{$T(W)$ for XCS (or XRS):} Here, within a given 2D weight matrix of a DNN layer, there are chunks of successive zero weight vectors of the size of crossbar-column (or crossbar-row) (see Fig. \ref{xbar}(b)-Bottom) which are eliminated.
\vspace{-3mm}
\end{enumerate}

Note, for standard unpruned DNNs, no transformation $T(W)$ is needed. The  resulting transformed weight matrices are then partitioned into  multiple  crossbar  instances. The next stage (functional modelling) of the platform converts  the  weights in the crossbars  to  suitable  conductances $G$ (between $G_{MIN}$ and $G_{MAX}$).  Thereafter,  the circuit-level non-idealities  are integrated  as parasitic resistances via  circuit laws  (Kirchoff’s laws and Ohm’s law) and  linear  algebraic operations  written  in  Python \cite{jain2020rxnn, bhattacharjee2021efficiency}. Further,  the variations/non-linearities in the synapses are included with Gaussian profiling. In this work, we follow a \textit{device agnostic} approach for analyzing the impact of intrinsic circuit-level and synaptic crossbar non-idealities on DNNs during inference. The various synapse parameters (e.g., $R_{MIN}$, $R_{MAX}$, device ON/OFF ratio) and values of the non-idealities used for our experiments have been listed in the table shown in Fig. \ref{hardware_eval}. The non-ideal synaptic conductances $G'$ are  then  converted back into non-ideal weight values and finally, for each DNN layer, we recombine all the crossbars and apply the inverse transformation $T^{-1}$ to obtain 2D matrices of non-ideal weights ($W'$). $W'$s are  integrated into  the original Pytorch based DNN model to conduct inference.

\vspace{-4mm}


\section{Experiments}
\label{sec:expt}

\begin{table}[t]
\caption{Table showing software accuracies and crossbar-compression-rates (with 32$\times$32 crossbars) for the various DNN models with CIFAR10 and CIFAR100 datasets}
\label{acc}
\resizebox{\linewidth}{!}{%
\begin{tabular}{|c|c|c|c|c|}
\hline
\textbf{Dataset:   CIFAR10} & \multicolumn{4}{c|}{\textbf{Software Accuracy (\%) $\parallel$ \textcolor{red}{Crossbar-compression-rate}}}               \\ \hline
\textbf{Network} & \textbf{Unpruned} & \textbf{C/F ($s=0.8$)} & \textbf{XCS ($s=0.8$)} & \textbf{XRS ($s=0.8$)} \\ \hline
\textbf{VGG11}              & 83.6 $\parallel$ \textcolor{red}{--}                 & 83.5 $\parallel$ \textcolor{red}{19.69$\times$}         & 83.28 $\parallel$ \textcolor{red}{4.26$\times$}         & 82.67 $\parallel$ \textcolor{red}{4.88$\times$}      \\ \hline
\textbf{VGG16}              & 84.48  $\parallel$ \textcolor{red}{--}              & 83.65 $\parallel$ \textcolor{red}{19.60$\times$}        & 82.06  $\parallel$ \textcolor{red}{5.57$\times$}       & 83.47 $\parallel$ \textcolor{red}{4.89$\times$}      \\ \hline
\textbf{Dataset: CIFAR100}  & \multicolumn{4}{c|}{\textbf{Software Accuracy (\%)}}               \\ \hline
\textbf{Network}            & \textbf{Unpruned}    & \multicolumn{3}{c|}{\textbf{C/F ($s=0.6$)}} \\ \hline
\textbf{VGG11}              & 53.29 $\parallel$ \textcolor{red}{--}               & \multicolumn{3}{c|}{52.72 $\parallel$ \textcolor{red}{5.64$\times$}}                  \\ \hline
\textbf{VGG16}              & 51.83 $\parallel$ \textcolor{red}{--}             & \multicolumn{3}{c|}{50.55 $\parallel$ \textcolor{red}{4.20$\times$}}                  \\ \hline
\end{tabular}%
}
\vspace{-6mm}
\end{table}

In this work, we train VGG11 and VGG16 DNNs with structured sparsity (via C/F pruning, XCS or XRS) using benchmark datasets such as, CIFAR10 and CIFAR100. For the experiments with CIFAR10 dataset, the sparsity is set as $s = 0.8$, while with CIFAR100 dataset, the sparsity is $s = 0.6$. The unpruned and pruned DNN models are trained to have nearly equal software accuracies to conduct a fair comparison of the impact of non-idealities when the models are mapped onto non-ideal crossbars (see Table \ref{acc}). The crossbar-compression-rates for the pruned DNNs on 32$\times$32 crossbars are also shown in Table \ref{acc}.
\vspace{-4mm}

\section{Impact of Non-idealities on Pruned DNNs}
\label{sec:init_res}

\begin{figure*}[t]
    \centering
    \includegraphics[width=.9\linewidth]{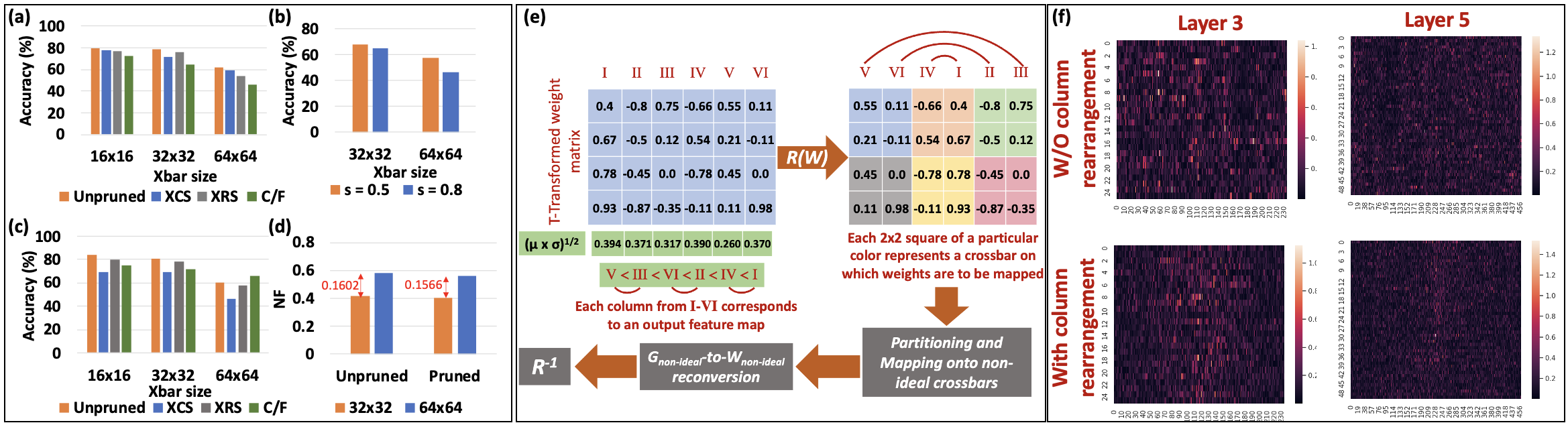}%
    \caption{Plot of inference accuracy versus crossbar size for- (a) unpruned and structure-pruned ($s=0.8$) VGG11/CIFAR10 DNN; (b) different values of sparsity ($s$) of a C/F pruned VGG11/CIFAR10 DNN; (c) unpruned and structure-pruned ($s=0.8$) VGG16/CIFAR10 DNN; (d) Plot showing the variation in average NF for unpruned and C/F pruned weight matrices on increasing the crossbar size from 32$\times$32 to 64$\times$64; (e) Pictorial representation of $R$ transformation integrated with our hardware evaluation framework. Note, in our hardware evaluation framework, when $R$ transformation is applied before partitioning and mapping the weights onto non-ideal crossbars, an inverse transformation $R^{-1}$ is applied after the integration of crossbar non-idealities to correctly carry out inference with the DNN model; (f) Heatmaps to visualize the impact of $R$ transfromation on the weight matrices of $3^{rd}$ \& $5^{th}$ layers of the VGG16/CIFAR10 DNN trained with C/F pruning with $s=0.8$}
    \label{initial_res}
 \vspace{-5mm}   
\end{figure*}

\begin{figure*}[t]
    \centering
    \includegraphics[width=.9\linewidth]{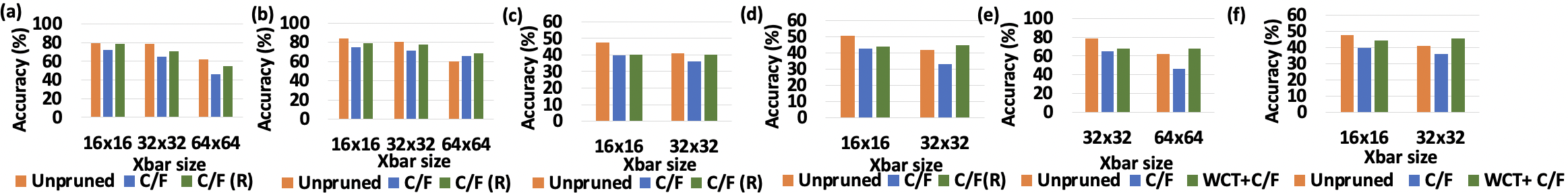}%
    \caption{A plot of inference accuracy versus crossbar size for unpruned, C/F pruned and- (a) C/F pruned with transformation $R$ ($s=0.8$) VGG11/CIFAR10 DNNs; (b) C/F pruned with transformation $R$ ($s=0.8$) VGG16/CIFAR10 DNNs; (c) C/F pruned with transformation $R$ ($s=0.6$) VGG11/CIFAR100 DNNs; (d) C/F pruned with transformation $R$ ($s=0.6$) VGG16/CIFAR100 DNNs; (e) WCT+C/F pruned ($s=0.8$) VGG11/CIFAR10 DNNs; (f) WCT+C/F pruned ($s=0.6$) VGG11/CIFAR100 DNNs}
    \label{mitigation_res}
    \vspace{-7mm}
\end{figure*}
 
The sparse DNNs have greatly reduced number of parameters than their unpruned counterparts which results in significantly lesser number of crossbars on hardware (see Table \ref{acc}). However, the fewer parameters remaining in the sparse DNNs are crucial for the model's performance. Thus, any non-ideality interfering with the fewer parameters of the sparse DNNs would have huge impact on the performance. In Fig. \ref{initial_res}(a), we find that for the VGG11/CIFAR10 model, the DNNs with structured sparsity (via C/F pruning, XCS, XRS with $s=0.8$) suffer greater accuracy degradation than their unpruned counterparts for crossbar sizes ranging from 16$\times$16 to 64$\times$64. Further, as we increase the crossbar size, both the accuracies of unpruned and pruned networks decline owing to increase in crossbar non-idealities \cite{bhattacharjee2021efficiency, chakraborty2020geniex}. Specifically, on 64$\times$64 crossbars, the inference accuracy of the unpruned model reduces by $\sim21\%$ with respect to the software baseline while, for the sparse DNNs pruned via C/F pruning, XCS and XRS, the decline is $\sim39\%$, $\sim24\%$ and $\sim30\%$, respectively. Also, in Fig. \ref{initial_res}(b), we find that on reducing the extent of sparsity in the C/F pruned DNNs from $s=0.8$ to $s=0.5$, the performance degradation suffered by the pruned DNNs is reduced. This validates the fact that greater sparsity, although leads to energy- and area-efficient mappings on crossbars, increases the interference of crossbar non-idealities, thereby hampering the performance of the pruned networks.

In Fig. \ref{initial_res}(c), for the VGG16 DNN with CIFAR10 dataset, the trends are similar to the case of the VGG11 DNN for XCS, XRS and C/F pruning ($s = 0.8$) in case of 16$\times$16 and 32$\times$32 crossbars. However, in case of a larger 64$\times$64 crossbar, we find that the performance of the network pruned by C/F pruning exceeds that of the unpruned network. This is because unpruned DNNs require a larger absolute number of crossbars for mapping than pruned ones. As a result, the value of NF is expected to increase at a higher rate for unpruned DNN on moving from 32$\times$32 to 64$\times$64 crossbars (see Fig. \ref{initial_res}(d)). So, for larger crossbars, the accuracy degradation for structure-pruned DNNs would decelerate compared to their unpruned counterparts, which can even lead to better absolute accuracy of the pruned networks than the unpruned ones. 

\vspace{-4mm}
\section{Non-Ideality Mitigation Strategies \& Results}
\label{sec:mitigation}

\subsection{Crossbar-Column rearrangement ($R$)}
\label{rearr}

For the sparse DNNs obtained via C/F pruning, we propose a simple hardware-friendly transformation of column rearrangement $R$ after the transformation $T$ (as discussed in Section \ref{sec:method}) before partitioning and mapping weights onto non-ideal crossbars. \textit{This transformation is inspired from the fact that the impact of non-idealities (or non-ideality factor NF) reduces for crossbars with higher proportion of low conductance synapses} \cite{bhattacharjee2021switchx, bhattacharjee2021neat}. Additionally, this approach of column rearrangement does not have any training overhead and is applied during the mapping of the DNNs onto crossbars. 

Let us consider a 4$\times$6 weight matrix $W$, after applying the transformation $T$, to be mapped onto 2$\times$2 crossbars (see Fig. \ref{initial_res}(e)). During the $R$ transformation, we first compute the value of $(\mu \times \sigma)^\frac{1}{2}$ for each column from I-VI, where $\mu$ and $\sigma$ respectively denote the mean and standard deviation of the absolute values of weights in each column. Thereafter, based on the increasing order of $(\mu \times \sigma)^\frac{1}{2}$, we rearrange columns I-VI in the manner shown. Now, in Fig. \ref{initial_res}(f), we visualize the impact of $R$ transformation on the weight matrices of the $3^{rd}$ and $5^{th}$ convolutional layers of the VGG16/CIFAR10 DNN (C/F pruned with $s=0.8$) using heatmaps. Before applying the transformation, the lighter (low conductance synapse) and darker (high conductance synapse) points in the heatmaps are intermixed. Post transformation, the lighter points are concentrated at the center of the heatmaps and darker points are mostly near the peripheries. Thus, post $R$ transformation and partitioning, individual crossbars have greater proportions of low conductance synapses, thereby mitigating the impact of crossbar non-idealities. 

\textbf{CIFAR10 \& CIFAR100 Results: } Fig. \ref{mitigation_res}(a-b) \& Fig. \ref{mitigation_res}(c-d) show that $R$ transformation improves the performance of the C/F pruned VGG11 and VGG16 DNNs. Specifically, $\sim9\%$ ($\sim6\%$)  improvement in accuracy is observed for VGG11 (VGG16) DNN on 64$\times$64 (32$\times$32) crossbars with CIFAR10 dataset. We also find that on 32$\times$32 crossbars, the accuracy of the pruned VGG16/CIFAR100 DNN post $R$ transformation is $\sim3\%$ greater than the unpruned counterpart. 

\vspace{-5mm}

\subsection{Weight-Constrained-Training (WCT)}


We introduce \textit{Weight-Constrained-Training} (WCT) of the structure-pruned DNN models on software before mapping onto crossbars, motivated by a recent work called Non-linearity Aware Training (NEAT) \cite{bhattacharjee2021neat}. In WCT, based on the weight distributions of all the layers of a trained DNN, we heuristically determine a cut-off value $W_{cut}$ for its weights, and then apply the following transformation on the weights of the DNN: $W = min\{|W|, W_{cut})\}*sign(W)$. This transformation constrains the DNN weights in the interval [$-W_{cut}$, $W_{cut}$]. With the above transformation, the DNN is iteratively trained for 2 epochs, to maintain nearly iso-accuracy with baseline. Note, the iterative training via WCT does not add any computational overhead to the overall training time, thereby making it a viable choice. The WCT-DNN results in greater proportion of low conductance states on the crossbars, thus, reducing the impact of crossbar non-idealities. The resultant sparse WCT-DNNs are then mapped onto crossbars. In Fig. \ref{mitigation_res}(e-f), we find that the WCT-DNNs maintain their performance even on increasing the crossbar size, making them resilient against crossbar non-idealities. Further, WCT-DNNs have better accuracy than the C/F pruned DNNs on crossbars. Specifically, with CIFAR10 (CIFAR100) dataset, the WCT-DNN has $\sim6\%$ ($\sim7\%$) higher accuracy than the unpruned model on 64$\times$64 (32$\times$32) crossbars.

\vspace{-4mm}

\section{Conclusion}
\label{sec:conclusion}

This work elucidates that DNNs with high degree of structured sparsity can lead to poor performance on non-ideal analog crossbars, albeit being more hardware-efficient. In our study, we have considered structured-pruning techniques, such as C/F pruning, XCS and XRS, and have shown that increased sparsity leads to loss in inference accuracy of the pruned DNNs on crossbars (considering non-idealities) with respect to their unpruned counterparts. To mitigate the non-ideality induced performance degradation, we introduce two hardware-centric strategies (crossbar-column rearrangement and WCT) that can be integrated with the mapping of the DNNs onto non-ideal crossbars to reduce the impact of non-idealities, thereby enhancing the performance of the sparse DNNs on crossbars. 

\vspace{-3mm}

\footnotesize
\section*{Acknowledgement}
This work was supported in part by C-BRIC, a JUMP center sponsored by DARPA and SRC, the National Science Foundation (Grant\#1947826) and the DARPA AI Exploration (AIE) program.
\normalsize

\vspace{-4mm}
\bibliographystyle{IEEEtran}

\bibliography{reference}

\end{document}